\documentclass[conference]{IEEEtran}
\IEEEoverridecommandlockouts
\usepackage{cite}
\usepackage{amsmath,amssymb,amsfonts}
\usepackage{algorithmic}
\usepackage{graphicx}
\usepackage{textcomp}
\usepackage{xcolor}
\usepackage{bm}
\usepackage{amstext}
\usepackage{amsfonts}
\usepackage{multirow}

\def\BibTeX{{\rm B\kern-.05em{\sc i\kern-.025em b}\kern-.08em
    T\kern-.1667em\lower.7ex\hbox{E}\kern-.125emX}}
\begin{document}

\title{Optimized latent-code selection for explainable conditional text-to-image GANs
}
\author{\IEEEauthorblockN{Zhenxing Zhang, Lambert Schomaker}
\IEEEauthorblockA{\textit{Bernoulli Institute, University of Groningen }\\
Groningen, The Netherlands \\
z.zhang@rug.nl, l.r.b.schomaker@rug.nl}
}

\maketitle

\begin{abstract}
The task of text-to-image generation has achieved remarkable progress due to the advances in the conditional generative adversarial networks (GANs). However, existing conditional text-to-image GANs approaches mostly concentrate on improving both image quality and semantic relevance but ignore the explainability of the model which plays a vital role in real-world applications. In this paper, we present a variety of techniques to take a deep look into the latent space and semantic space of the conditional text-to-image GANs model. We introduce pairwise linear interpolation of latent codes and `linguistic' linear interpolation to study what the model has learned within the latent space and `linguistic' embeddings. 
Subsequently, we extend linear interpolation to triangular interpolation conditioned on three corners to further analyze the model.
After that, we build a ${Good}$/${Bad}$ data set containing unsuccessfully and successfully synthetic samples and corresponding latent codes for the image-quality research. Based on this data set, we propose a framework for finding ${good}$ latent codes by utilizing a linear SVM. 
 Experimental results on the recent DiverGAN generator trained on two benchmark data sets qualitatively prove the effectiveness of our presented techniques, with a better than 94\% accuracy in predicting ${Good}$/${Bad}$ classes for latent vectors. The ${Good}$/${Bad}$ data set is publicly available at https://zenodo.org/record/5850224$\#$.YeGMwP7MKUk. 
\end{abstract}

\section{Introduction}
The task of text-to-image generation aims to automatically produce visually realistic and semantically correlated samples according to the given natural-language descriptions. Research in conditional generative adversarial networks (cGANs) and style generative adversarial networks (GANs) has rapidly attracted much attention in recent years, since these can be potentially applied in various fields, such as art creation, computer-aid design, data augmentation for training image classifiers, photo-editing, the education of young children, etc. Existing text-to-image generation methods \cite{zhang2021dtgan, zhang2018stackgan++, zhu2019dm, tao2020df} have achieved tremendous improvements in both image quality and semantic consistency with the cGANs, mapping a latent code into an image sample semantically aligning with the input text via adversarial training. Despite the promising results, it is still difficult to explain what a conditional text-to-image GANs model has learned within the latent space and semantic space. How to understand the relation between the textual (linguistic) probes and the generated image properties? This constitutes the first research topic of the current contribution.

One particular disadvantage of synthetic image-generation algorithms is that the performance evaluation is more difficult than is the case in classification problems where a `hard' accuracy can be computed. In case of cGANs this problem is most clearly present for end users: How to ensure that generated images are believable, realistic or natural? In current literature, the good examples are often cherry picked while occasionally also the less successful samples are shown. However, for actual use in data augmentation or in artistic applications, one would like to guarantee that generated images are good, i.e., of a sufficiently believable natural quality. Given the high dimensionality of latent codes, there is a very high prior probability of non-successful patterns to be generated for a given input noise probe. How to construct a random latent-code generator with an increased probability of drawing successful samples? After the generator/discriminator pair has done its best effort, apparently additional constraints are necessary.


To address these difficult but highly relevant questions, we take a closer look at the latent space and `linguistic' embeddings of the conditional text-to-image GANs architecture using linear interpolation between latent codes for textual contrasts, triangular interpolation and Good/Bad classification on generated images.

We start with a well-trained generator for text-to-image synthesis. For the first question, we qualitatively analyze the roles played by latent vectors and `linguistic' embeddings in the synthetic-image semantic space through pairwise {\em linear interpolation} of latent codes and {\em linear interpolation} between keywords. We show that although semantic properties contained in the picture change continuously in the latent space, the appearance of the image does not always vary smoothly along with the contrasting word embeddings. In addition, we extend linear interpolation to {\em triangular interpolation} for simultaneously investigating three latent codes or three keywords in the give textual description. 

To tackle the second question, we need an evaluation method to separate `good' latent vectors from `bad' latent codes. This is done by constructing a ${Good}$/${Bad}$ data set comprising 210 ${Good}$ samples and 210 ${Bad}$ samples, and corresponding latent vectors. These samples are generated by the recent DiverGAN \cite{zhang2021divergan} generator pre-trained on the CUB bird data set \cite{wah2011caltech}.  With the help of the ${Good}$/${Bad}$ data set, we can develop a technique distinguishing ${Good}$ pictures (${Good}$ latent codes) from ${Bad}$ samples (${Bad}$ latent codes). Bengio et al. \cite{bengio2013better} postulate that deep convolutional networks have the ability to linearize the manifold of pictures into a Euclidean subspace of deep features. Inspired by this hypothesis, we argue that ${Good}$ samples and ${Bad}$ samples can be classified by a approximately linear boundary in such deep-feature space. Therefore, we propose a framework to discover ${Good}$ latent codes by adopting a linear SVM. These Good latent codes can be utilized for further research including data augmentation and latent-space direction discovery \cite{shen2020interpreting, wold1987principal}.


We carry out comprehensive experiments on the recent DiverGAN \cite{zhang2021divergan} generator trained on two widely used data sets (i.e., the CUB bird \cite{wah2011caltech} and MS COCO \cite{lin2014microsoft} data sets), since DiverGAN has the ability to employ a generator/discriminator pair to synthesize diverse and high-quality samples. The experimental results in the current study represent an improvement in performance and explainability in the analyzed algorithm \cite{zhang2021divergan}. The contributions of this work can be summarized as follows:

$\bullet$ We introduce pairwise linear interpolation of latent codes and linear interpolation between contrastive keywords for an improved explainability of the conditional text-to-image GANs model. 

$\bullet$ We extend linear interpolation to triangular interpolation for further analysis using a 2-simplex, i.e., triplet contrast, and better data augmentation.  

$\bullet$ We build a ${Good}$/${Bad}$ data set to study how to ensure the quality of synthetic samples while presenting a framework to discover ${Good}$ latent vectors. 


\section{Related Work}
\subsection{Text-to-Image Generation}
Thanks to recent unprecedented advances in generative approaches especially GANs and cGANs, existing text-to-image generation methods have achieved tremendous progress in both image quality and semantic consistency.

Reed et al. \cite{10.5555/3045390.3045503} was the first to utilize the cGANs to generate visually realistic samples according to detailed textual descriptions.
Zhang et al. \cite{zhang2017stackgan, zhang2018stackgan++} proposed to exploit a multi-stage pipeline to yield high-quality images from natural-language descriptions in a coarse-to-fine way. Qiao et al. \cite{qiao2019mirrorgan} developed MirrorGAN exploiting a semantic-preserving text-to-image-to-text framework, where an image-caption model was customized to improve the semantic consistency between image contents and textual descriptions. Zhu et al. \cite{zhu2019dm} presented DMGAN which adopted a dynamic memory module to boost the quality of initial image samples. Li et al. \cite{li2019object} built ObjGAN that paid attention to salient-object generation in complex scene. CPGAN \cite{liang2020cpgan} designed a memory structure to parse the produced image in an object-wise manner and introduced a conditional discriminator to promote the semantic alignment of text-image pairs. Tao et al. \cite{tao2020df} designed a matching-aware zero-centered gradient penalty (MA-GP) loss addressing the issues of the multi-stage framework. Zhang et al. \cite{zhang2021dtgan} presented DTGAN leveraging two new attention models and conditional adaptive instance-layer normalization to produce perceptually realistic samples with a generator/discriminator pair. Zhang et al. \cite{zhang2021divergan} proposed DiverGAN which can yield diverse and visually plausible samples which are semantically correlated with given natural-language descriptions. Note that our experiments are conducted on the DiverGAN generator due to its superior performance in both image quality and diversity.
\subsection{Latent Space Manipulation}
Understanding the latent space of the GANs plays a significant role in controlling the image-generation process. Recently, a series of supervised and unsupervised approaches have been presented for identifying semantically meaningful latent-space directions of the well-trained GANs. The attributes of a synthetic sample can be edited by moving the corresponding latent vector along the acquired directions. 

Shen et al. \cite{shen2020interpreting} presented InterfaceGAN employing label data (e.g., smile and pose) to learn different SVM decision boundaries as useful directions. Goetschalckx et al. \cite{goetschalckx2019ganalyze} developed a pipeline, GANalyze, to discover interpretable directions as high-level cognitive attributes with the help of an assessor model (e.g., MemNet \cite{khosla2015understanding}). Voynov et al. \cite{voynov2020unsupervised} proposed to find, in an unsupervised way, latent-space directions based on a matrix and a classifier. Härkönen et al. \cite{harkonen2020ganspace} presented GANSpace which collected a set of latent codes and conducted PCA \cite{wold1987principal} on them to obtain principal components as primary directions in the latent space. Wang et al. \cite{wang2021hijack} introduced an iterative scheme to gain semantic control over the image-generation process. Shen et al. \cite{shen2021closed} proposed a closed-form factorization algorithm directly decomposing the pre-trained weights for semantic image editing. 

\section{The Proposed Approach}
\label{sec:3}
In this section, we discuss pairwise linear interpolation of latent codes and `linguistic' linear interpolation for an improved explainability of the conditional text-to-image GANs frameworks. Afterwards, triangular interpolation conditioned on three corners is introduced to further study the model. Subsequently, we describe the proposed method for discovering $Good$ latent codes. 
\subsection{Preliminary}\label{sec:3.1}
The goal of text-to-image generation is, given textual descriptions, to automatically produce perceptually realistic image samples. Let $S=\{ (I_{i}, C_{i})\}_{i=1}^{N}$ represent a set of $N$ image-text pairs for training, where $I_{i}$ indicates an image and $C_{i}=(c_{i}^{1}, c_{i}^{2}, ..., c_{i}^{K})$ denotes a suite of $K$ natural-language descriptions, while $S$ is cast into a training set and a testing set. Current conditional text-to-image GANs \cite{zhang2018stackgan++, zhu2019dm, zhang2021dtgan} approaches commonly follow the same paradigm. The generator $G$ aims at yielding a visually plausible and semantically consistent sample $\hat{I}_{i}$ 
according to a latent code randomly sampled from a fixed distribution and a text description $c_{i}$ randomly picked from $C_{i}$, where $c_{i}= (w_{1}, w_{2}, ..., w_{m})$ contains $m$ words. The discriminator $D$ of the GANs is trained to distinguish the real image-text pair $(I_{i}, c_{i})$ from the synthetic image-text pair $(\hat{I}_{i}, c_{i})$. 
Through the above training process, a text-to-image generation model is assumed to have the ability to synthesize diverse and semantically related pictures, given different textual descriptions and injected noise. Despite high-quality pictures achieved by the existing methods, we yet do not understand what a text-to-image synthesis architecture has learned within the latent-vector space and semantic space, nor is it known how to construct a successful sample that would be accepted by human users. In this paper, we introduce a variety of basic techniques to provide insights into the explainability of the text-to-image synthesis framework.

In order to understand `embeddings' or latent codes in deep learning, several methods have been proposed. A common method is to visualize the space using, e.g., t-SNE or k-means clustering. This may give some insight on the location of dominant image categories in the sub space.
An alternative approach is to utilize - yet another - step of dimensionality reduction by applying standard PCA on the embedding. However, this still does not lead to good explanations and an easy controllability of the image-generation process. In this paper, we propose to start from instances of successfully generated images and analyze intermediate patterns, i.e., latent codes that are positioned between two successful samples.
\subsection{Linear Interpolation}
We start with a well-trained and fixed generator $G(z, (w, s))$, which takes a random noise $z$, word embeddings $w$ of the given text description and the corresponding sentence vector $s$ as input and outputs a photo-realistic and semantically consistent sample. We introduce a linear-interpolation technique for qualitatively analyzing the roles played by random latent vectors and textual descriptions in synthesizing samples.

\noindent\textbf{Pairwise linear interpolation of latent codes.} To better understand how $G(z, (w, s))$ utilizes the latent space to achieve diversity, we first visualize the samples generated by linear interpolation between a given successful starting-point latent code $z_{0}$ and a successful end-point latent code $z_{1}$ . We can see whether the appearance of the synthetic samples will change continuously along the interpolated latent-vector trajectory, since visually smooth linear interpolations are often regarded as a marker of the performance of a generative model \cite{michelis2021linear}. Specifically, pairwise linear interpolation of latent codes is defined as:
\begin{equation}
 f(\gamma)=G((1-\gamma)z_{0}+\gamma z_{1}, (w, s)) \ \ for \ \ \gamma \in [0,1]
 \label{e1}
\end{equation}
where $\gamma$ is a scalar mixing parameter.

As an example, for the CUB bird data set, we assume that both the background and the visual appearance of birds proper (the foreground) will vary gradually with the variations of the latent codes. This would imply that semantics contained in the picture also change continuously in the latent space. More importantly, when given a single natural-language description, the quality of generated images depends on injected random noise but should change smoothly along with the latent vectors, which we detail in Section \ref{sec:3.4}.

\noindent\textbf{Linear interpolation and semantic interpretability.} In addition to pairwise linear interpolation of latent codes, we also study linear interpolation between keywords focusing on a qualitative exploration how well $G(z, (w, s))$ exploits the `linguistic' space as well as testing the influence of individual, different words on the generated sample. We can observe how the samples vary as a word in the given text is replaced with another word, for instance by using a polarity axis of qualifier key words (dark-light, red-blue, ...).
To be specific, we can first acquire two word embeddings (i.e., $w_{0}$ and $w_{1}$) and two corresponding sentence vectors (i.e., $s_{0}$ and $s_{1}$) by only altering a significant word (e.g., the color attribute value and the background value) in the input natural-language description. Afterwards, the results are obtained by performing linear interpolation between the initial textual description $ (w_{0}, s_{0})$ and the changed description $ (w_{1}, s_{1})$ while keeping the latent code $z$ fixed. Mathematically, this proposed `linguistic' linear interpolation combines the latent code, the word and the sentence embeddings and is formulated as:
\begin{equation}
h(\gamma)=G(z, (1-\gamma)w_{0}+\gamma w_{1}, (1-\gamma)s_{0}+\gamma s_{1}) 
\end{equation}
where $\gamma \in [0,1]$ is a scalar mixing parameter and $z$ is a successful latent code.

For the CUB bird dataset, when we vary the color attribute value in the given sentence, we empirically explore what happens in the color mix: Do we, e.g., get an average color interpolation in RGB space or does the network find another solution for the intermedate points between two disparate embeddings (Section \ref{4.2})? 

In general, our presented `linguistic' linear interpolation has the following advantages: 

$\bullet$~Linear interpolation between keywords can be utilized to quantitatively control the attribute of the synthetic sample, when the attribute varies smoothly with the variations of the word vectors. For example, the length of the beak of a bird can be adjusted precisely via `linguistic' linear interpolation between the word embeddings of `short' and `long'. 

$\bullet$~When the attribute of the synthetic sample does not change gradually along with the word embeddings, we can exploit `linguistic' linear interpolation to produce a variety of novel samples. Take bird synthesis as an example: When conducting linear interpolation between color keywords, $G(z, (w, s))$ is likely to generate a new bird whose body contains two colors (e.g., red patches and blue patches) in the middle of the interpolation results, as shown in Fig~\ref{fig3}.

$\bullet$~Through linear interpolation between keywords, we can take a deep look into which keywords play important roles in yielding foreground images as well as which image (background) regions are determined by the linguistic terms. 

\subsection{Triangular Interpolation}\label{sec:3.2}
We extend linear interpolation between two points to an interpolation between three points, i.e., in the 2-simplex, for further studying $G(z, (w, s))$ and better performing data augmentation. Since this kind of interpolation forms a triangular plane, we name it triangular interpolation. Triangular interpolation is able to generate more and more diverse samples conditioned on three corners (e.g., latent vectors and keywords), spanning a field rather than a line.

\noindent\textbf{Triangular interpolation of latent codes.} Similar as in Equation~\ref{e1}, triangular interpolation of latent codes is achieved by employing three successful latent codes (i.e., $z_{0}$, $z_{1}$ and $z_{2}$) as corner points to perform an interpolation in the 2-simplex field. It is denoted as: 
\begin{equation}
f(\gamma_{1}, \gamma_{2})=G((1-\gamma_{1}-\gamma_{2})z_{0}+\gamma_{1} z_{1}+\gamma_{2} z_{2}, (w, s)) 
\end{equation}
where $\gamma_{1} \in [0,1]$ and $\gamma_{2} \in [0,1]$ are mixing scalar parameters and $z_{0}$ represents the initial latent code. When we fix the value of $\gamma_{2}$, triangular interpolation between $z_{0}$, $z_{1}$ and $z_{2}$ can be viewed as pairwise linear interpolation between $z_{0}$ and $z_{1}$. Additionally, the number of samples synthesized by linear interpolation will decrease as $\gamma_{2}$ increases. Therefore, when we sample 10 points with the initial interpolation (i.e., $\gamma_{2}=0$), we can obtain 55 pictures via triangular interpolation of latent codes.
We want to explore how semantic properties contained in the images also change with the variations of the latent vectors for triangular interpolation of latent codes. Some linguistic contrast may represent a smooth transition, whereas others will involve clear transition points.
As an example, the question may be asked whether the midpoint between a blue and red corner is represented by purple in RGB, or by a patchy transition involving red and blue image elements flowing into one another. 

\noindent\textbf{Triangular interpolation and semantic interpretability.} Similar to linear interpolation between keywords, we need to derive three word embeddings (i.e., $w_{0}$, $w_{1}$ and $w_{2}$) and three corresponding sentence vectors (i.e., $s_{0}$, $s_{1}$ and $s_{2}$) as corners to define the presented `linguistic' triangular interpolation:
\begin{equation}
\begin{split}
h(\gamma_{1}, \gamma_{2})=G(z, &(1-\gamma_{1}-\gamma_{2})w_{0}+\gamma_{1} w_{1}+\gamma_{2} w_{2}, \\
&(1-\gamma_{1}-\gamma_{2})s_{0}+\gamma_{1} s_{1}+\gamma_{2} s_{2})
\end{split}
\end{equation}
where $\gamma_{1} \in [0,1]$ and $\gamma_{2} \in [0,1]$ are mixing scalar parameters and $z$ is a successful latent vector. 

For the sake of attribute analysis, we can obtain three new textual descriptions by replacing the attribute word in the initial natural-language description with another two attribute words. Then, through triangular interpolation between keywords, the generator has the ability to yield pictures based on the above three attributes. Moreover, we expect that `linguistic' triangular interpolation should achieve the same visual smoothness as `linguistic' linear interpolation. In other words, when fixing the weight (i.e., $\gamma_{2}$) of the third text in  triangular interpolation between keywords, the attributes of the image vary gradually along with the word embeddings if the interpolation results of `linguistic' linear interpolation between the first two textual descriptions change continuously.

The `linguistic' triangular interpolation has obvious advantages over  linear interpolation between keywords. Firstly, `linguistic' triangular interpolation is able to produce more images to do data augmentation than linear interpolation. Secondly, we can concurrently control two different attributes (e.g., color and the length of the beak) via  triangular interpolation between keywords. Thirdly, through `linguistic' triangular interpolation, three identical attributes (e.g., red, yellow and blue) can be combined to synthesize a novel sample.
\begin{figure}[t]
  \begin{minipage}[b]{1.0\linewidth}
  \centerline{\includegraphics[width=85mm]{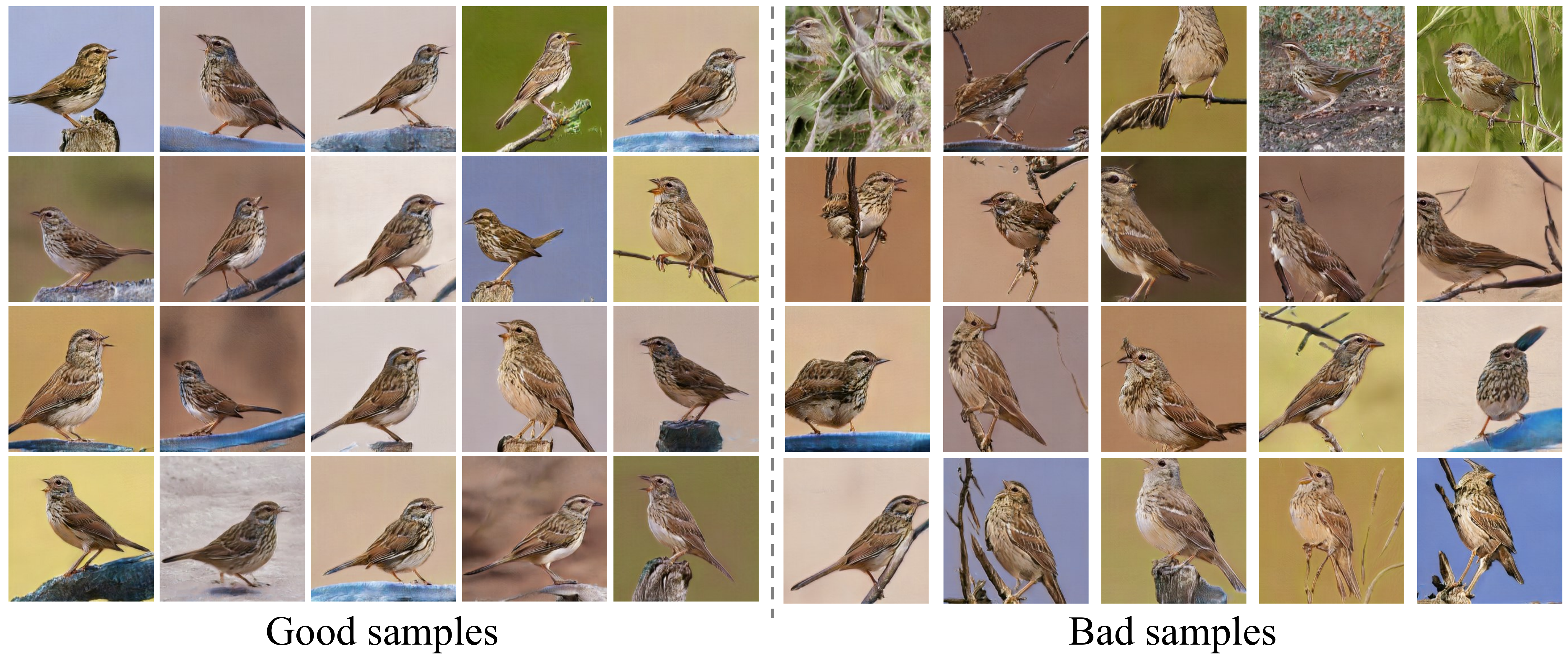}}
  \end{minipage}
  \caption{A snapshot of the ${Good}$/${Bad}$ data set: the \textbf{left} column is from the ${Good}$ data set; the \textbf{right} column is from the ${Bad}$ data set.}
  \vspace{-0.1in}
  \label{fig7} 
\end{figure}
\begin{figure*}[t]
   \begin{minipage}[b]{1.0\linewidth}
   \centerline{\includegraphics[width=180mm]{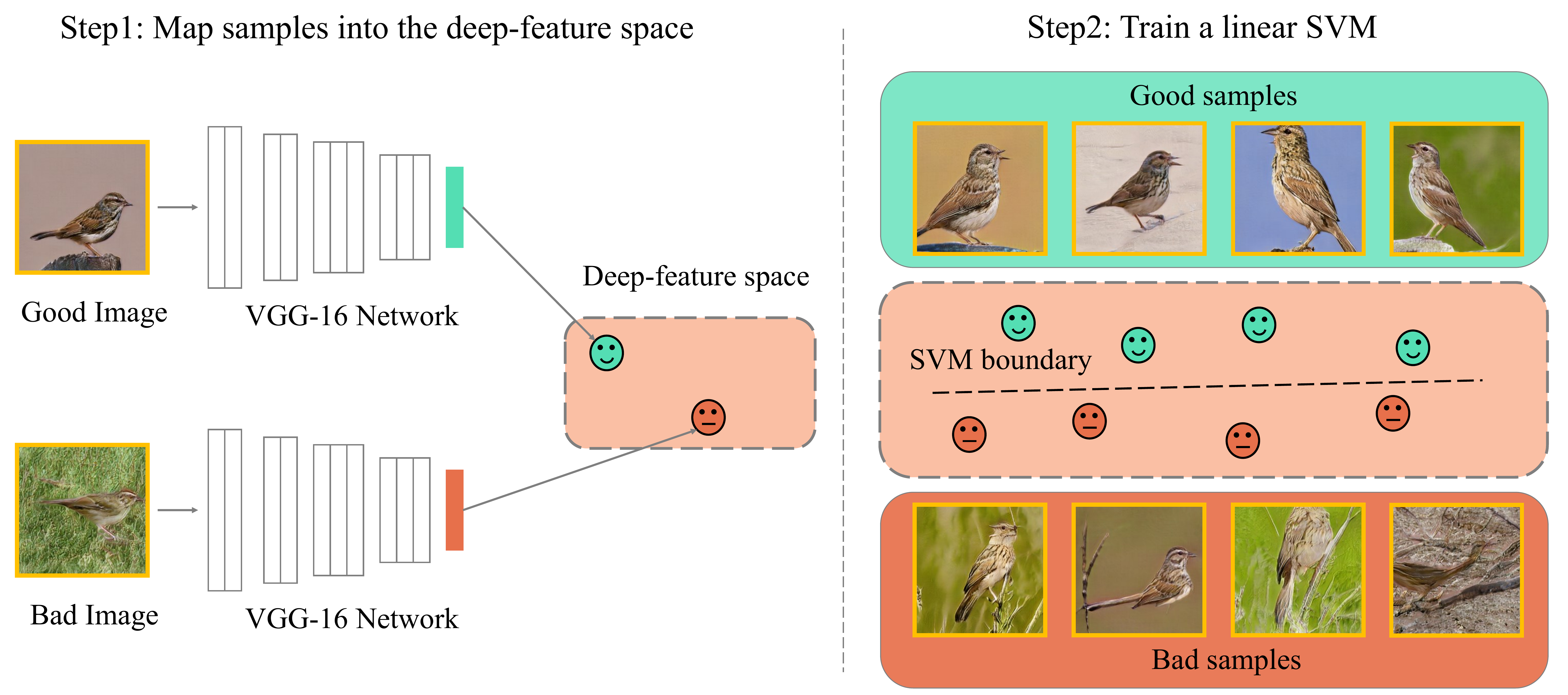}}
   \end{minipage}
   \caption{A schematic outline of the first two steps for finding ${Good}$ latent codes.}
   \vspace{-0.1in}
  \label{fig1} 
\end{figure*}

\subsection{Finding Good Latent Codes}\label{sec:3.4}
When the samples conditioned on the first and the last latent vectors are visually realistic, we empirically discover that pairwise  linear  interpolation  of  latent  codes is likely to yield a series of high-quality images. We name these latent vectors ${Good}$ latent codes and these pictures ${Good}$ samples. At the same time, we find that interpolation results often look blurry if the generator $G(z, (w, s))$ cannot generate plausible samples according to these two latent codes called ${Bad}$ latent codes. Moreover, when the start latent vector is ${Good}$ but the final latent vector is ${Bad}$, the first part of interpolation results is usually realistic but the final part is not plausible. 

Therefore, we assume that $G(z, (w, s))$ may synthesize high-quality pictures based on the latent codes around a ${Good}$ latent vector. In addition, we hypothesize that there may be a non-linear boundary between ${Good}$ samples and ${Bad}$ samples. In this subsection, we introduce a simple framework for finding ${Good}$ latent codes. More importantly, when we obtain these ${Good}$ latent vectors, we can synthesize large numbers of satisfactory results, e.g., for the purpose of data augmentation. 

In order to verify our hypothesis, we build a ${Good}$/${Bad}$ data set consisting of 420 synthetic samples (i.e., 210 ${Good}$ birds and 210 ${Bad}$ birds) which are divided into a training set (i.e., 150 ${Good}$ birds and 150 ${Bad}$ birds) and a testing set (i.e., 60 ${Good}$ birds and 60 ${Bad}$ birds). These samples are generated by the DiverGAN generator trained on the CUB bird data set \cite{wah2011caltech}. We visualize a snapshot of our data set in Fig \ref{fig7}. The details of collecting the ${Good}$/${Bad}$ data set will be described in Section \ref{4.4}.

It is difficult to directly apply a traditional classifier (e.g., a linear SVM) to separate realistic images adequately from inadequate samples or distinguish ${Good}$ latent codes from ${Bad}$ latent vectors, since the image instances exist in a non-linear manifold \cite{weinberger2006unsupervised} and latent codes are drawn from a fixed distribution (e.g., the standard normal distribution). 
In the meantime, we cannot train a deep neural network (e.g., VGG \cite{simonyan2014very}) from scratch to label a synthetic sample as ${Good}$ or ${Bad}$ due to the small number of the samples in the ${Good}$/${Bad}$ training set. 
Bengio et al. \cite{bengio2013better} postulate that deep convolutional networks have the ability to linearize the manifold of pictures into a Euclidean subspace of deep features. Inspired by this hypothesis, we expect that ${Good}$ and ${Bad}$ samples can be classified by an approximately linear boundary in such deep-feature space. The process of discovering plausible samples and corresponding latent codes is summarized in three steps (depicted in Fig \ref{fig1}):

1. We employ the publicly available VGG-16 network trained on ImageNet \cite{russakovsky2015imagenet} to map the samples from the training set (i.e., 150 ${Good}$ samples and 150 ${Bad}$ samples) into the deep-feature representation;

2. We utilize the above deep features and the corresponding labels to fit a linear SVM labeling the samples in the deep-feature space;

3. The testing samples are mapped into deep features and predicted by the trained SVM. Then, we can derive ${Good}$ samples and the corresponding ${Good}$ latent vectors.

We expect that this procedure is highly effective, accurately predicting the labels of the samples from the testing set. As a result, we have the ability to automatically select realistic samples from the synthetic images while obtaining ${Good}$ latent codes. More importantly, a wealth of suitable samples can be produced by performing pairwise  linear  interpolation  of  latent  codes, triangular interpolation of latent  codes and other operations on these ${Good}$ latent vectors. 
\section{Experiments}
\label{er}
In this section, to evaluate the effectiveness of the presented techniques, we perform a set of experiments on the recent DiverGAN \cite{zhang2021divergan} generator trained on the CUB bird \cite{wah2011caltech} and MS COCO \cite{lin2014microsoft} datasets. To be specific, we clarify the details of experimental settings in Section \ref{4.1}. After that, the experiments in Section \ref{4.2} and Section \ref{4.3} are carried out to evaluate the linear-interpolation and triangular-interpolation techniques. Subsequently, the experiments in Section \ref{4.4} are performed to prove the effectiveness of the proposed $Good$ latent-code discovery method. 
\begin{figure}[t]
  \begin{minipage}[b]{1.0\linewidth}
  \centerline{\includegraphics[width=85mm]{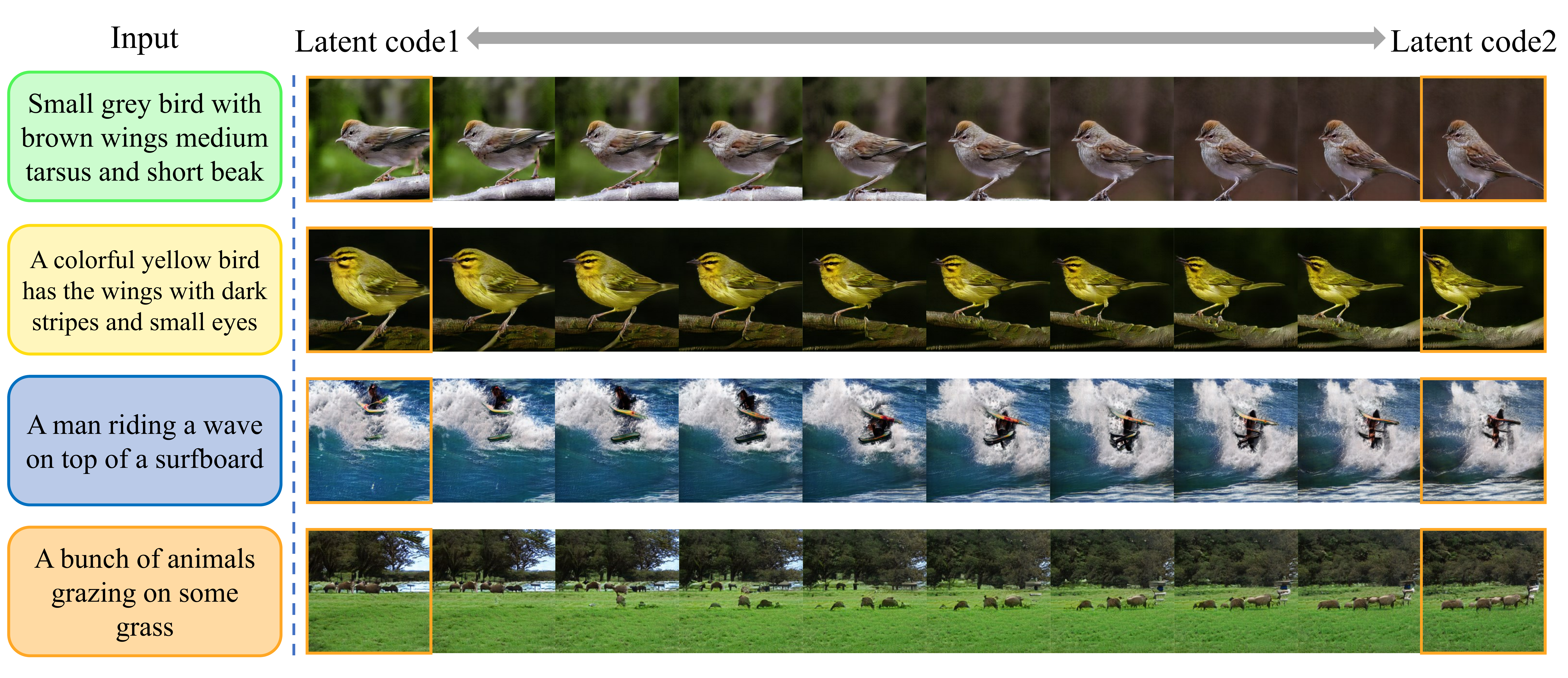}}
  \end{minipage}
  \caption{Pairwise interpolation of DiverGAN random latent-code samples on the CUB and COCO datasets, for four text input probes.}
  \vspace{-0.1in}
  \label{fig2} 
\end{figure}

\subsection{Experimental Settings}
\label{4.1}
\noindent\textbf{Datasets.} Our experiments are carried out on two types of extensively used text-to-image generation data sets: CUB bird and MS COCO data sets. The CUB bird data set contains 8,855 training images and 2,933 testing images. Each image has 10 corresponding natural-language descriptions. The MS COCO data set is a more challenging data set, consisting of 82,783 training images and 40,504 testing images. Each picture is accompanied by 5 human-annotated captions. 
\begin{figure*}[t]
  \begin{minipage}[b]{1.0\linewidth}
  \centerline{\includegraphics[width=180mm]{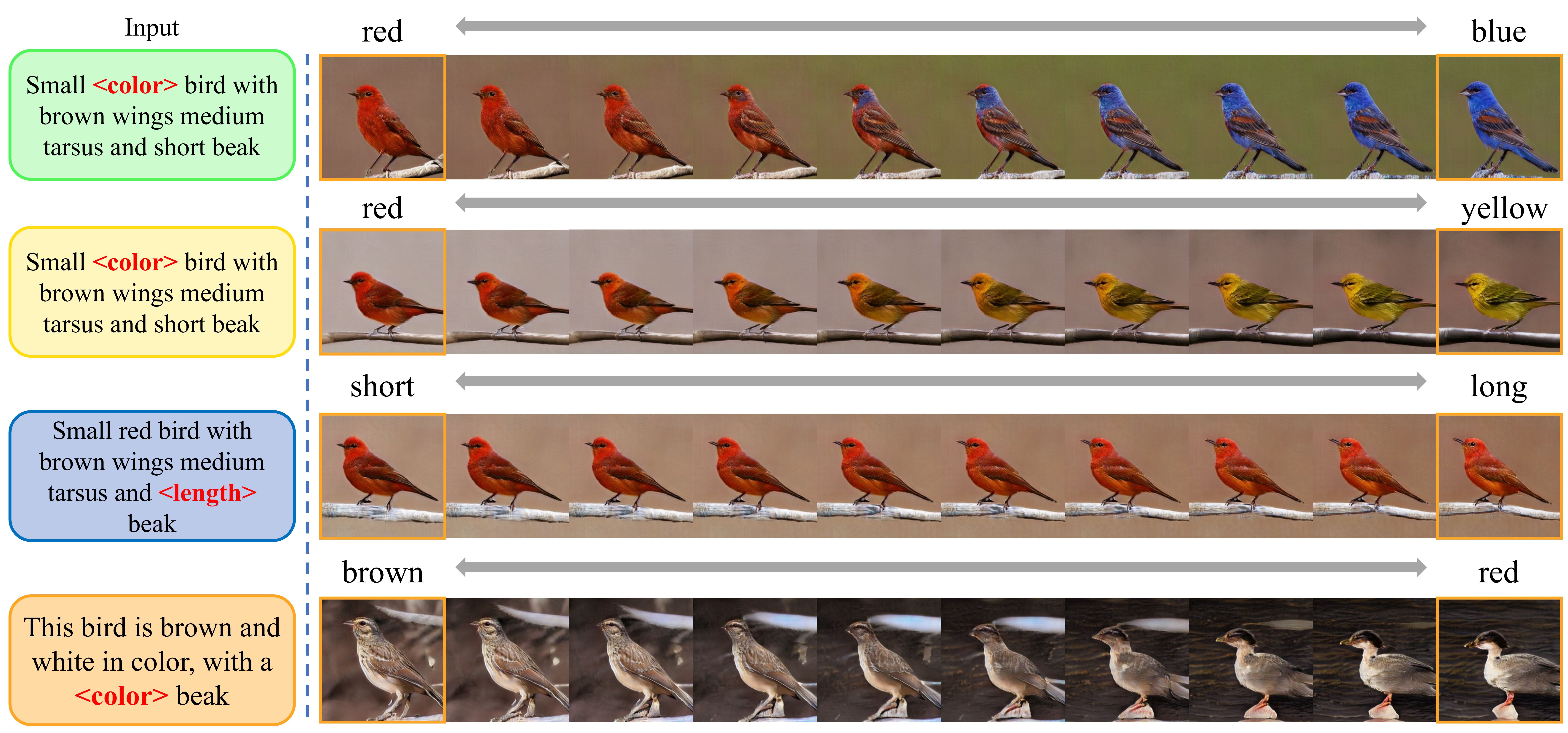}}
  \end{minipage}
  \caption{`Linguistic' interpolation of DiverGAN random latent-code samples on the CUB dataset, for four text input probes.}
  \vspace{-0.1in}
  \label{fig3} 
\end{figure*}
\begin{figure*}[t]
  \begin{minipage}[b]{1.0\linewidth}
  \centerline{\includegraphics[width=180mm]{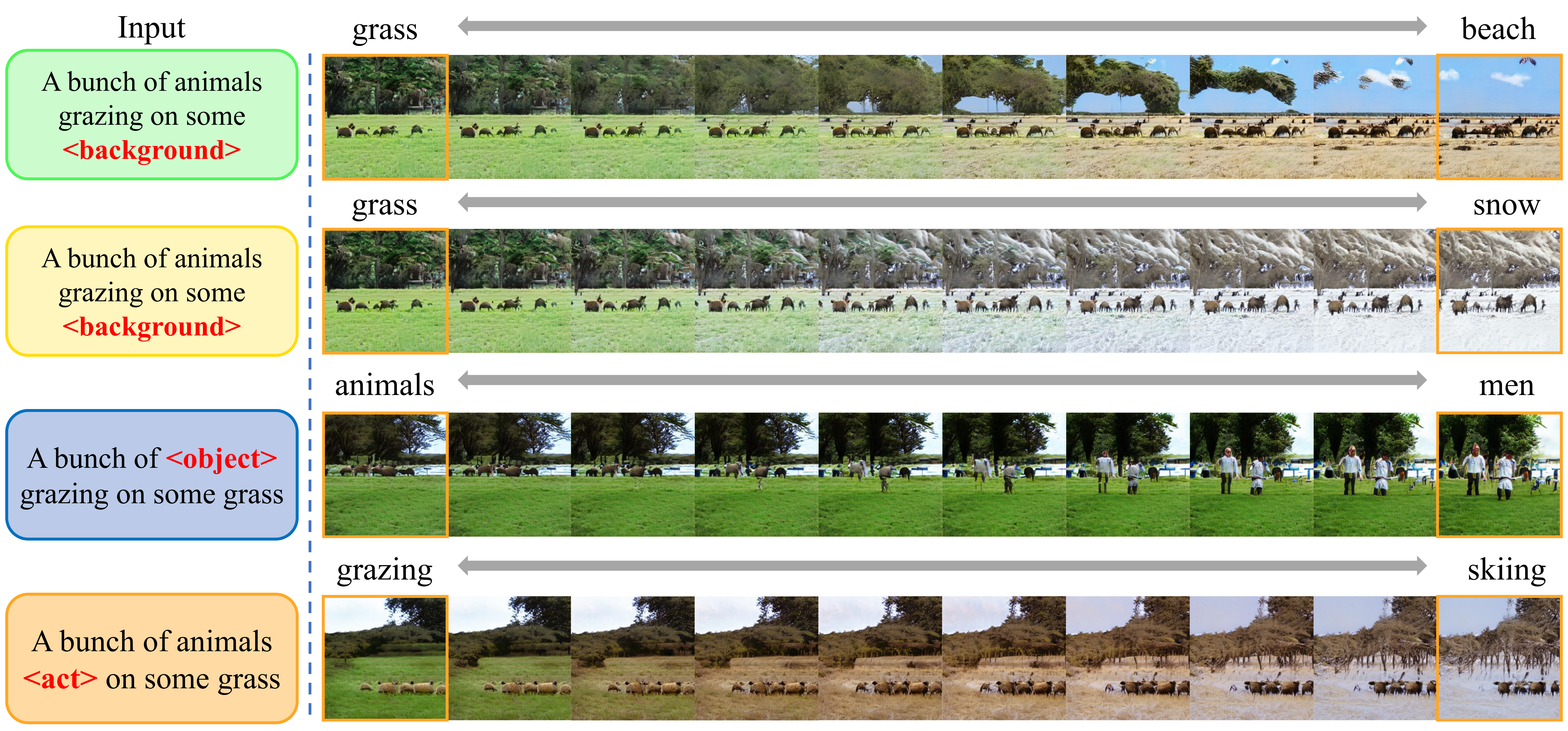}}
  \end{minipage}
  \caption{`Linguistic' interpolation of DiverGAN random latent-code samples on the COCO dataset, for four text input probes.}
  \vspace{-0.1in}
  \label{fig4} 
\end{figure*}

\noindent\textbf{Implementation details.} We take the recent DiverGAN generator as the backbone generator, which is pre-trained on the CUB bird and MS COCO data sets. The steps of linear interpolation are set to 10. We set the steps of $\gamma_{1}$ and $\gamma_{2}$ in triangular interpolation to 10. 
Our approach is implemented by PyTorch~\cite{paszke2019pytorch}. A single NVIDIA Tesla V100 GPU (32 GB memory) is employed for all experiments.  

\subsection{Results of Linear Interpolation}
\label{4.2}
In this section, we evaluate the linear-interpolation approach.

\noindent\textbf{Pairwise linear interpolation of latent codes.} The pairwise linear-interpolation results of DiverGAN are presented in Fig~\ref{fig2}. For the CUB bird data set, we can see that both the background and the visual appearances of footholds ($1^{st}$ row), both the shapes and the positions of birds ($2^{nd}$ row) change gradually with the variances of latent codes. For the COCO dataset, we can observe that both the visual appearances of the man and the ocean wave ($3^{rd}$ row), both the animals and the background ($4^{th}$ row) vary continuously along with the latent vectors. It suggests that semantic properties contained in the picture change smoothly in the latent space for DiverGAN. 
\begin{figure}[t]
  \begin{minipage}[b]{1.0\linewidth}
  \centerline{\includegraphics[width=85mm]{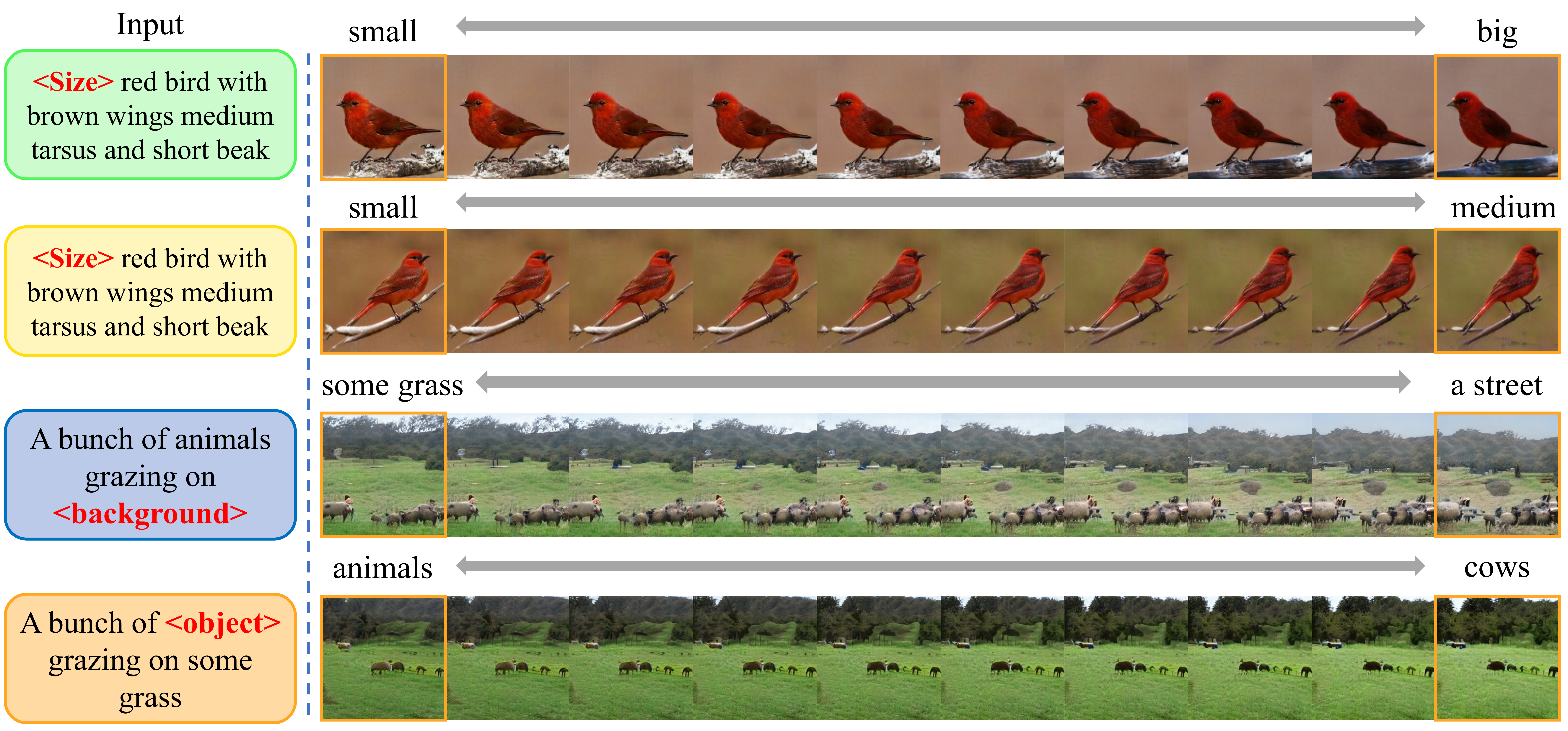}}
  \end{minipage}
  \caption{Unsuccessful `linguistic' interpolation of DiverGAN random latent-code samples on the CUB and COCO datasets, for four text input probes. For the third row, the desired attribute (i.e., a street) is not emerging.}
  \vspace{-0.1in}
  \label{fig5} 
\end{figure}

\noindent\textbf{`Linguistic' linear interpolation.} Fig~\ref{fig3} shows the qualitative results of `linguistic’ linear interpolation of DiverGAN on the CUB bird data set, indicating that the attributes correlated with the synthetic sample do not always change gradually with the variations of word embeddings. For instance, the color of the bird does not vary continuously from `red’ to `blue’ in the first row. In the medium of interpolation results, DiverGAN generates multiple novel birds, whose bodies are composed of red and blue patches. However, the color attribute of the bird changes gradually from ‘red’ to ‘yellow’ in the second row. We are able to acquire an average color interpolation in RGB space by merging the first and second attributes.
We can also see that in the third row, the length of the beak varies smoothly along with textual vectors while other attributes remain unchanged. Furthermore, while the color of the beak changes continuously with the variations of word embeddings, the shape of the bird varies largely in the fourth row. The above results suggest that DiverGAN has the ability to capture the significant words (e.g., the color of the body and the length of the beak) in the given textual description. More importantly, by exploiting the characteristic as well as linear interpolation between keywords, we can precisely control the image-generation process while producing various novel samples. 

The qualitative results of linear interpolation between keywords on the COCO data set are shown in Fig~\ref{fig4}. We can observe that DiverGAN accurately identifies `beach’, `snow’ and `men’ while generating the corresponding image samples. In addition, the background ($1^{st}$ and $2^{nd}$ row) and the object ($3^{rd}$ row) change continuously along with `linguistic’ vectors.
It can also be seen that although we change the `acting' word from `grazing’ to `skiing’, the background significantly varies from `grass’ to `snow’ in the fourth row, which demonstrates that some words (e.g., `skiing’) play a vital role in the generation process of image samples. The above analysis also indicates that when given adequate training images, DiverGAN is able to control the background (e.g., from grass to beach) and the object (e.g., from animals to men) of complex scenes with the help of `linguistic’ linear interpolation, since DiverGAN is able to learn the corresponding semantics in the `linguistic’ space. 
\begin{figure}[t]
  \begin{minipage}[b]{1.0\linewidth}
  \centerline{\includegraphics[width=85mm]{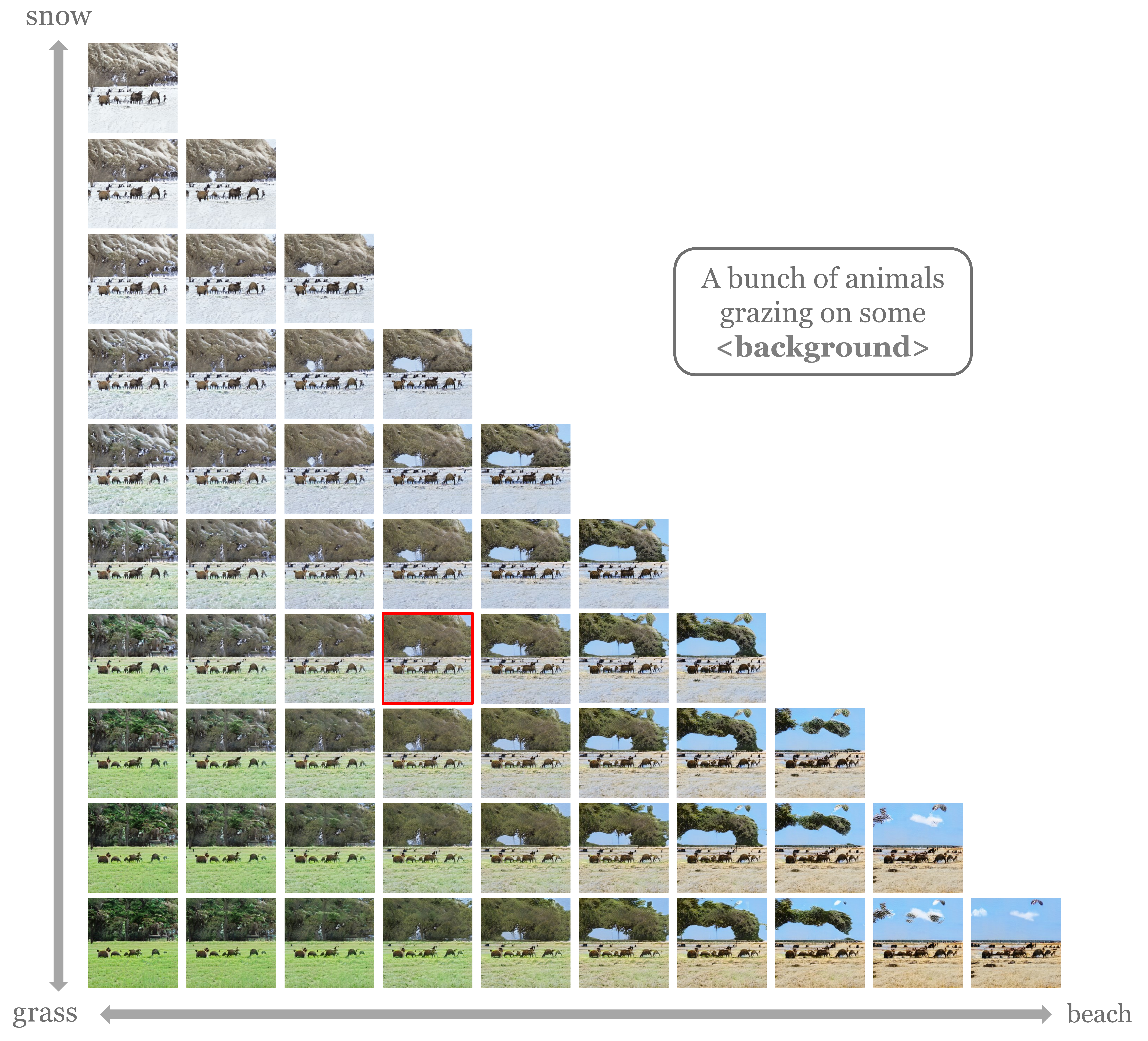}}
  \end{minipage}
  \caption{Triangular interpolation of latent codes, for linguistic attributes ${snow,grass,beach}$ on two dimensions. The center is marked in red.}
  \label{fig6} 
\end{figure}

In addition to visualizing effective examples of linear interpolation between keywords, we also present some unsuccessful results in Fig~\ref{fig5}. As can be observed in Fig~\ref{fig5}, the size of the bird ($1^{st}$ and $2^{nd}$ row) does not vary with the variations of the word (from `small’ to `big’ and from `small’ to `medium’). We can also see that the background ($3^{rd}$ row) and the object ($4^{th}$ row) unfortunately do not change along with the word (from `grass’ to `street’ and from `animals’ to `cows’). At this point we can conclude that many meaningful contrasts can be learned (Fig~\ref{fig4}), but there are areas where the method is not able to capture important variations along a dimension. This may be due to architectural or data-related limitations. In order to improve our insights, we will look at triangular interpolation in the next section.

\subsection{Results of `Linguistic' Triangular Interpolation}
\label{4.3}
The triangular interpolation for linguistic attributes (i.e., the points between ${snow,grass,beach}$) in two dimensions is shown in Fig~\ref{fig6}. We can observe that the transitions towards the three corner points are natural as well as smooth. Furthermore, the interpolation results achieve a balanced triangular shape within the triangle, such that the center marked in red is the combination of three linguistic attributes. If the application concerns data augmentation, 55 believable samples are obtained by performing triangular interpolation between keywords. 
\begin{table}
\begin{center}
\begin{tabular}{c c}
\hline
Methods & Accuracy($\%$) \\
\hline
Image& 70.0 \\
PCA-Image & 73.3 \\
Latent Code & 75.8 \\
VGG-16(conv5$\_$3)& 94.2\\
VGG-16(conv5$\_$2)& 96.7\\
\hline
$\textbf{VGG-16(conv5$\_$1)}$& $\textbf{97.5}$\\
\hline
\end{tabular}
\end{center}
\caption{Classification accuracy on the separation boundary with respect to image quality. {\em Image} refers to a direct application of SVM on the image pixels. {\em PCA-Image} refers to using PCA on the image pixels after reducing the dimensionality to 128 and applying SVM to identify realistic samples. {\em Latent Code} refers to the direct application of SVM in the latent space.}
\label{tab:1}
\end{table}
\begin{figure*}[t]
  \begin{minipage}[b]{1.0\linewidth}
  \centerline{\includegraphics[width=180mm]{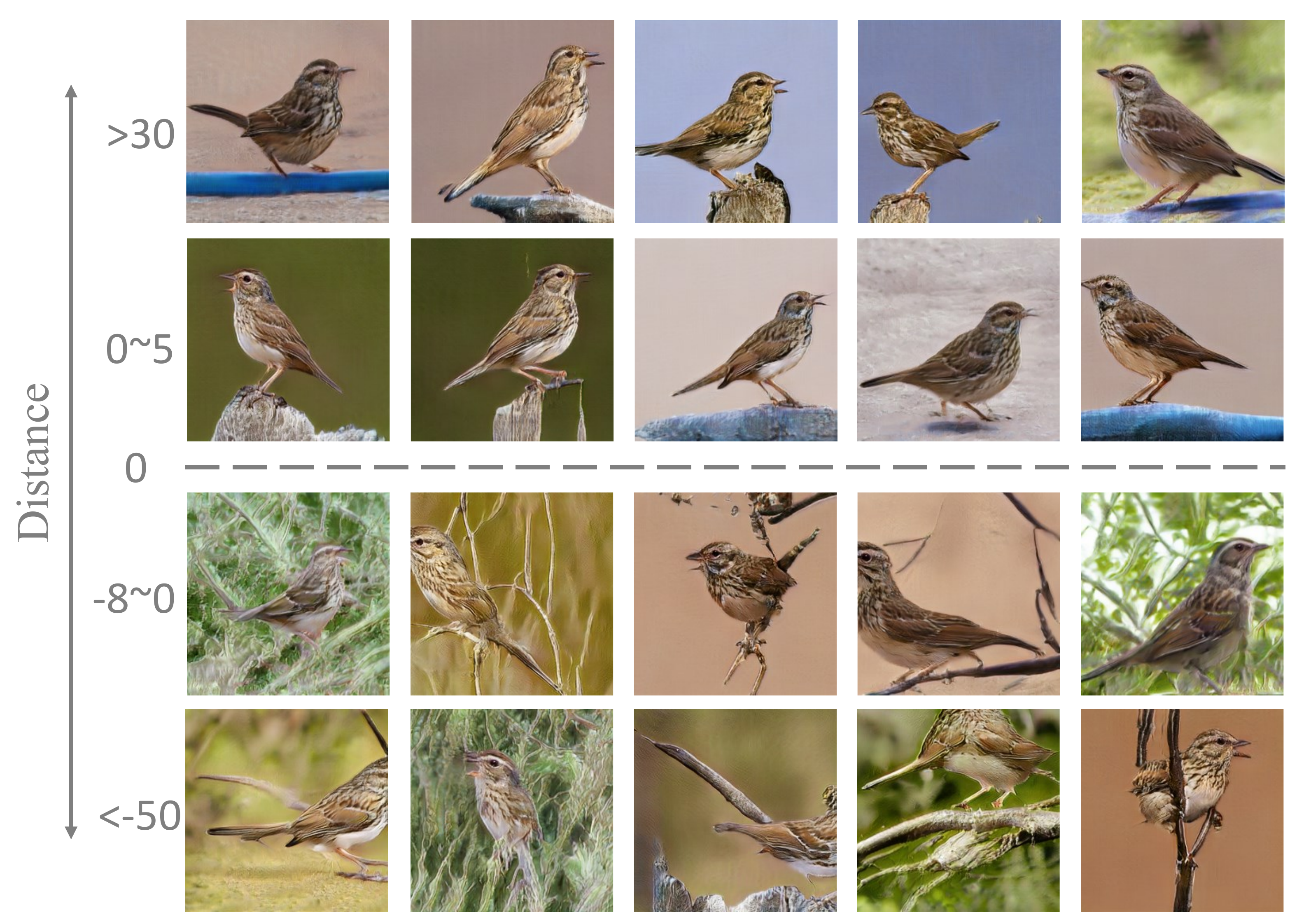}}
  \end{minipage}
  \caption{Example of partitioning of latent-code space between ${Good}$ (two top rows) and ${Bad}$ latent codes (two bottom rows), as determined by the discriminant value (distance) computed by a linear SVM (training set: Ngood=150, Nbad=150.)}
  \vspace{-0.1in}
  \label{fig8} 
\end{figure*}

\subsection{Results of Finding Good Latent Codes}
\label{4.4}
We describe here the details of building the ${Good}$/${Bad}$ data set. The first stage of the construction of the ${Good}$/${Bad}$ data set involves exploiting DiverGAN pre-trained on the CUB bird data set to yield a large number of candidate samples. Then, we label the synthetic pictures as ${Good}$ (extremely realistic), ${Bad}$ (very blurry) and uncertain. Afterwards, 150 ${Good}$ birds and 150 ${Bad}$ birds are leveraged as the training set while 120 images (i.e., 60 ${Good}$ birds and 60 ${Bad}$ birds) are utilized to evaluate the model.

We adopt different methods to discover ${Good}$ latent vectors. The results are reported in Table~\ref{tab:1}. Here, we discover that all methods using the learned feature vectors of a well-trained VGG-16 network achieve over 94$\%$, suggesting that there exists a (almost) linear boundary in the deep-feature space which can accurately distinguish ${Good}$ images from ${Bad}$ samples. In addition, the conv5\_1 activation in the pre-trained network obtains the best performance (accuracy: 97.5$\%$). We also attempted to employ the SVM with radial basis function (RBF) kernel to classify deep features, acquiring the same result as the linear SVM. Moreover, it can be observed that directly operating on the image pixels (accuracy: 70.0$\%$) and the latent space (accuracy: 75.8$\%$) does not work well for the classification of ${Good}$ and ${Bad}$ samples/latent codes. To boost the accuracy, we conduct PCA on the image pixels to reduce the dimension to 128 and apply a linear SVM to identify realistic samples. However, the accuracy is only improved by 3.3$\%$. The above results confirm the effectiveness of our proposed framework. Therefore, we are able to automatically discover ${Good}$ latent vectors.  More importantly, a suite of visually plausible pictures can be generated by performing linear interpolation, triangular interpolation and other operations on these ${Good}$ latent codes. 

We visualize some typical output samples selected from the test set (Ngood=60, Nbad=60) in Fig~\ref{fig8} according to their distance to the decision boundary of the trained SVM. It can be observed that ${Good}$ samples are distinguishable from ${Bad}$ samples. Meanwhile, the ${Bad}$ birds around the boundary may have higher quality than the ${Bad}$ birds far from the decision boundary. It should be noted that in non-ergodic problems, where there is not a natural single signal source for the good (or the bad) images, but there rather exists a partitioning of space, the SVM discriminant value for a sample is not guaranteed to be consistent with the intuitive prototypicality of the heterogeneous underlying class~\cite{van2014separability} due to the lack of a central density for that class.
\section{Conclusion}
In this paper, we present multiple techniques for an improved explainability of the conditional text-to-image GANs model. We evaluate our proposed approaches on the recent DiverGAN generator trained on two very different data sets, including the single-object CUB bird data set and the complex-scene COCO data set. We show that semantic properties contained in the image change gradually with the variations of latent codes, but the attributes of the sample do not always vary continuously along with the word embeddings. We also find that the generator of DiverGAN cannot capture the size of the object due to the mechanism of the convolutional neural network and cannot understand some words in the given textual description owing to the limitation of the data set. Simultaneously, experimental results show that the presented ${Good}$ latent-code discovery method achieves 97.5$\%$ accuracy on the testing data set, suggesting that image quality may be identified and ${Good}$ latent codes can be acquired by our approach. In the future, we plan to explore how to identify interpretable latent-space directions for the text-to-image generation framework. 



{\small
\bibliographystyle{IEEEtran}
\bibliography{egbib}
}

\end{document}